\documentclass[runningheads]{llncs}
\usepackage{latexsym}
\usepackage{booktabs}
\usepackage{graphicx}
\usepackage{amsmath}

\usepackage{comment}
\usepackage{multirow}
\usepackage{tikz}
\def\checkmark{\tikz\fill[scale=0.4](0,.35) -- (.25,0) -- (1,.7) -- (.25,.15) -- cycle;} 

\begin{document}
\title{Simplifying Paragraph-level Question Generation via Transformer Language Models}
\titlerunning{Simplifying Paragraph-level Question Generation}
%

\author{Luis Enrico Lopez\inst{1}* \and 
Diane Kathryn Cruz\inst{1}* \and
Jan Christian Blaise Cruz\inst{1}* \and
Charibeth Cheng\inst{1}}
\authorrunning{E. Lopez et al.}
\institute{De La Salle University Manila, Taft Ave., Malate, 1004 Manila, Philippines \\
\email{\{luis\_lopez,diane\_cruz,jan\_christian\_cruz,charibeth.cheng\}@dlsu.edu.ph}}


\maketitle              

\graphicspath{{img/}}

\begin{abstract}
Question Generation (QG) is an important task in Natural Language Processing (NLP) that involves generating questions automatically when given a context paragraph. While many techniques exist for the task of QG, they employ complex model architectures, extensive features, and additional mechanisms to boost model performance. In this work, we show that transformer-based finetuning techniques can be used to create robust question generation systems using only a single pretrained language model, without the use of additional mechanisms, answer metadata, and extensive features. Our best model outperforms previous more complex RNN-based Seq2Seq models, with an 8.62 and a 14.27 increase in METEOR and ROUGE\_L scores, respectively. We show that it also performs on par with Seq2Seq models that employ answer-awareness and other special mechanisms, despite being only a single-model system. We analyze how various factors affect the model's performance, such as input data formatting, the length of the context paragraphs, and the use of answer-awareness. Lastly, we also look into the model's failure modes and identify possible reasons why the model fails.

\end{abstract}

\section{Introduction}
{\let\thefootnote\relax\footnote{{*: Equal contribution. Order determined by drawing lots.}}}

Question Generation (QG) \cite{rus2008question}, while not as prominent as its sibling task Question Answering (QA), still remains a relevant task in NLP. The ability to ask meaningful questions provides evidence towards \textit{comprehension} within an Artificial Intelligence (AI) model \cite{nappi2017importance}. This makes the task of QG important in the bigger picture of AI.

Many studies have produced robust models with good performance for QG in recent years. The most widely-used techniques are Deep Learning-based approaches involving Sequence-to-Sequence (Seq2Seq) \cite{sutskever2014sequence} models. These approaches use two LSTM-based \cite{Hochreiter:1997:LSM:1246443.1246450} neural networks, one to encode the source context paragraph, and the other to decode the embedded information and output a generated question \cite{duan2017question}.

Further works that improve on the standard Seq2Seq-based QG models use either extra mechanisms, extra features, or both. These include the usage of extra linguistic features \cite{zhou2017Neural} or the introduction of answer-awareness \cite{zhao2018paragraph,du2017learning,dong2019unified}, which uses the answer to the desired question, or the position of the answer within the context paragraph as additional features. A combination of these techniques provide the base for state-of-the-art QG in recent years.

More recently, other techniques have been proposed in order to perform QG. Reinforcement Learning (RL) have produced consistent results for the task by using policy gradients \cite{yuan2017machine}. The use of Transformers \cite{vaswani2017attention} over standard RNNs have also been adopted as these models provide the power of Attention in order to refer to specific points of context within the context paragraph, alleviating the RNN's memory bottleneck \cite{dong2019unified}.

While all of these techniques are robust, they all employ complex models, extra features, and additional mechanisms that make them harder to train and expensive to reproduce. In this work, we show that transformer-based finetuning techniques can be used to create robust question generation systems using only a single pretrained language model, without the use of additional mechanisms, answer metadata, and extensive features. 

We show that our method, albeit simpler, produces results on par with the state-of-the-art. We benchmark standard language model finetuning on a reformatting of the SQuAD \cite{rajpurkar-etal-2016-squad} v.1.1 dataset and evaluate generation performance with standard language generation metrics. In addition, we perform a variety of analyses in order to isolate performance indicators within our model and identify its weaknesses and failure modes.

\section{Methodology}

\subsection{Data Preparation}
\label{sec-method-data}

\begin{figure*}[ht]
    \centering
    \includegraphics[width=12cm]{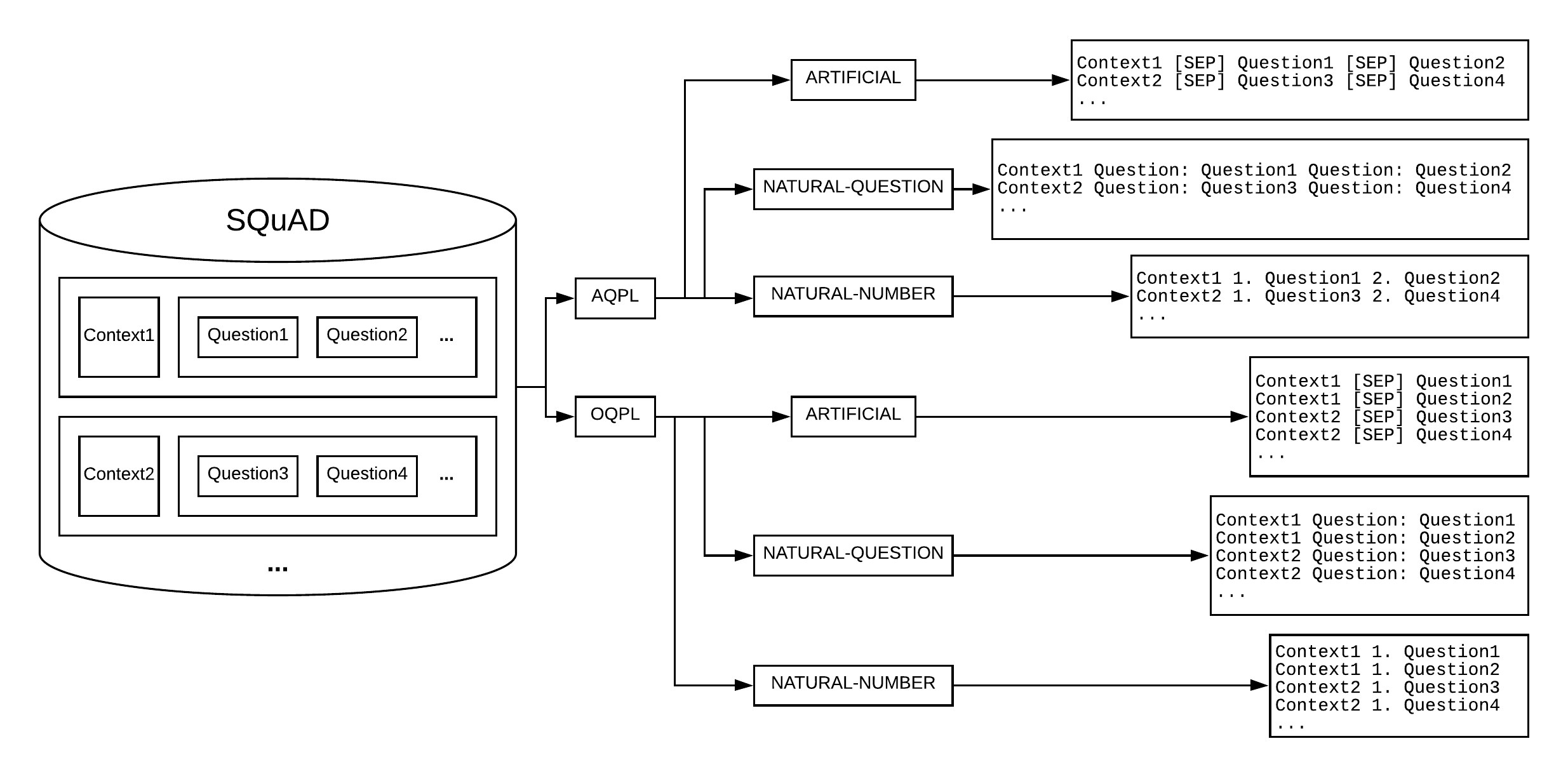}
    \caption{Data preparation pipeline for SQuAD.}
    \label{fig-data-prep-flow}
\end{figure*}

\begin{figure}[ht]
    \scriptsize
    \centering
    \begin{tabular}{|p{.9\linewidth}|}
        \hline
        \textcolor{red}{Super Bowl 50 was an American football game to determine the champion of the National Football League (NFL) for the 2015 season. The American Football Conference (AFC) champion Denver Broncos defeated the National Football Conference (NFC) champion Carolina Panthers 24–10 to earn their third Super Bowl title. The game was played on February 7, 2016, at Levi's Stadium in the San Francisco Bay Area at Santa Clara, California... } \textcolor[rgb]{0,0.5,0}{\texttt{[SEP]}} \textcolor{blue}{Which NFL team represented the AFC at Super Bowl 50?} \\
        \hline
    \end{tabular}
    \caption{A sample training example for question generation training. The context, delimiter, and question are highlighted in red, green, and blue respectively. Uses the ARTIFICIAL delimiter and the OQPL format. Text adapted from SQuAD. \protect{\cite{rajpurkar-etal-2016-squad}}.}
    \label{fig-sample-oqpl}
\end{figure}

\begin{figure}[ht]
    \scriptsize
    \centering
    \begin{tabular}{|p{.9\linewidth}|}
        \hline
        \textcolor{red}{Super Bowl 50 was an American football game to determine the champion of the National Football League (NFL) for the 2015 season. The American Football Conference (AFC) champion Denver Broncos defeated the National Football Conference (NFC) champion Carolina Panthers 24–10 to earn their third Super Bowl title. The game was played on February 7, 2016, at Levi's Stadium in the San Francisco Bay Area at Santa Clara, California. As this was the 50th Super Bowl, the league emphasized the "golden anniversary" with various gold-themed initiatives, as well as temporarily suspending the tradition of naming each Super Bowl game with Roman numerals (under which the game would have been known as "Super Bowl L"), so that the logo could prominently feature the Arabic numerals 50.} \textcolor[rgb]{0,0.5,0}{\texttt{[SEP]}} \textcolor{blue}{Which NFL team represented the AFC at Super Bowl 50?} \textcolor[rgb]{0,0.5,0}{\texttt{[SEP]}} \textcolor{blue}{Where did Super Bowl 50 take place?} \textcolor[rgb]{0,0.5,0}{\texttt{[SEP]}} \textcolor{blue}{What color was used to emphasize the 50th anniversary of the Super Bowl?} \\
        \hline
    \end{tabular}
    \caption{A sample training example for question generation training. The context, delimiter, and questions are highlighted in red, green, and blue respectively. Uses the ARTIFICIAL delimiter and the AQPL format. Text adapted from SQuAD dataset \protect{\cite{rajpurkar-etal-2016-squad}}.}
    \label{fig-sample-aqpl}
\end{figure}

We train the question generation model on version 1.1 of the Stanford Question Answering Dataset (SQuAD) \cite{rajpurkar-etal-2016-squad}. SQuAD contains context paragraphs, each with sets of questions and corresponding answer spans related to the contents of these paragraphs; in total, SQuAD contains more than 100,000 crowdsourced questions. While originally intended for the task of question answering, previous works on question generation \cite{du2017learning,zhao2018paragraph} have repurposed SQuAD as a training and test dataset, designating the questions as the target output rather than the answer spans.

As GPT-2 was pretrained to perform language modeling, we finetune it in a way similar to how it was trained on language modeling. Thus, we format SQuAD such that it appears similar to input data for language modeling. The entire dataset is transformed into a continuous body of text. Each training example consists of a context paragraph and its associated question(s) transformed into a single continuous sequence with a delimiter in between. Training examples are separated by the newline character \texttt{\textbackslash n}. Figure \ref{fig-sample-oqpl} shows an example of a single training example in this form.

There can be multiple ways to perform this transformation from the dataset's original representation (JSON for SQuAD) to a continuous language modeling-ready text. We experiment with two factors in formatting this data: the delimiter used, and the representation method for multiple questions per context paragraph. Figure \ref{fig-data-prep-flow} illustrates the six data formats we use for model training.

\subsubsection{Delimiters}
\label{sec-method-data-delimiters}
During data preparation, a delimiter is placed between each input context paragraph and output question. During training, this delimiter allows the model to properly distinguish between the context and question, while during prediction, it can be used as a marker at the end of some input text to invoke question generation behavior in the model. We experiment with three different delimiting schemes: 1) \textbf{ARTIFICIAL}, or a delimiter in the form of the token \texttt{[SEP]}, 2) \textbf{NATURAL-QUESTION}, or a delimiter in the form of the word \texttt{Question}, and 3) \textbf{NATURAL-NUMBER}, or a delimiting scheme in the form of a numbered list, where each item is a question. 

The ARTIFICIAL delimiter was not present in the original model's vocabulary, and its weights are learned from scratch during the finetuning phase, while the NATURAL delimiting schemes rely on token weights already learning during the pretraining phase, thus making it possible for the model's pretrained knowledge to affect performance through these delimiters. Similar keywords have been shown to be effective in invoking certain pretrained model behaviors (e.g. \texttt{TL;DR:} for summarization), even in a zero-shot setting \cite{radford2019language}.

\subsubsection{Questions Per Line}
\label{sec-method-data-qpl}

There can be several possible questions associated with a single paragraph. We experiment with two ways to flatten this many-to-one relationship in the formatted data:

\paragraph{All Questions Per Line (AQPL)}

A single training example consists of a context paragraph with all of its associated questions placed immediately after it, separated from one another with the selected delimiter. While this avoids duplication of context and thus results in faster training time, it may potentially result in the model no longer being able to attend to earlier tokens as its context window moves further away from the beginning of the input paragraph.

This is critical in the question generation task, as information pertaining to a reference question may be found anywhere in the input paragraph. If that information is found at the beginning, outside of the model's current context window, the model may have difficulty generating the corresponding question.

\paragraph{One Question Per Line (OQPL)}

Each context paragraph is duplicated for each of its associated questions, such that for a single training example, there is only one context and one question. For many cases, this may alleviate the moving context window problem raised with AQPL, as the length of a single training example is reduced to the length of an input paragraph plus the length of a single associated question. However, this format does result in a longer training time due to the duplicated contexts increasing the size of the final formatted dataset.

\subsection{Model Setup and Finetuning}
\label{sec-method-model}

For our base pretrained model, we used HuggingFace’s implementation \cite{Wolf2019HuggingFacesTS} of the 124 million parameter GPT-2, the smallest of the four available GPT-2 model sizes. From this base model, we finetuned six question generation models, each using one of the data format combinations enumerated in Section \ref{sec-method-data}.

We trained each model for 3 epochs using causal language modeling loss. We used the Adam optimizer \cite{DBLP:journals/corr/KingmaB14} with an initial learning rate of $5 \times 10^{-4}$ and a linearly decreasing learning rate schedule with warm up for 10\% of total training steps.

For training, we used a single Tesla V100 16GB GPU. As the model would not fit into memory using GPT-2's default maximum sequence length of 1024 and a batch size of 32, we simulated this batch size by combining an actual batch size of 2 with 16 gradient accumulation steps per minibatch.

We opted out of using the larger models because of time and hardware limitations; training the 345 million parameter GPT-2 with a single 16GB GPU would force us to use an actual batch size of 1 in order to fit the model into memory, greatly increasing training time, while training either of the two larger model sizes would require us to use multiple GPUs.

\subsection{Model Generation}

We set the model temperature to $0.6$. Higher temperature values result in more randomness in generations, while lower values approach greedy behavior.

We use the top-p nucleus sampling method \cite{holtzman-2019-top-p} with a value of $p = 0.9$. Top-p allows for more diverse generations than a purely greedy scheme, and minimizes the occurrence of certain tokens or token spans repeating indefinitely in the generated text.

Each generation loop is terminated either when the model generates the newline character \texttt{\textbackslash n}, or when the model reaches a generation length of 32 tokens. We manually set this maximum length in order to terminate generation sessions that are stuck in token span loops and do not reach the \texttt{\textbackslash n} end-of-text token on their own.

\subsection{Metrics and Evaluation}
\label{sec-method-eval}

Similar to the work of \cite{zhao2018paragraph}, we perform automatic evaluation metrics such as BLEU\_1, BLEU\_2, BLEU\_3, BLEU\_4 \cite{papineni-etal-2002-bleu}, ROUGE\_L \cite{lin-2004-rouge} and METEOR \cite{denkowski:lavie:meteor-wmt:2014}. We used the evaluation package made by \cite{sharma2017nlgeval} to quantify the models' performance.

\section{Results and Discussion}
\label{sec-results}

\begin{table*}[ht]
\centering
\begin{tabular}{llllllll}
\hline \textbf{Format} & \textbf{Delimiter} & \textbf{BLEU\_1} & \textbf{BLEU\_2} & \textbf{BLEU\_3} & \textbf{BLEU\_4} & \textbf{METEOR} & \textbf{ROUGE\_L} \\ \hline
\multirow{3}{4em}{AQPL} &
Artificial & 54.83 & 30.13 & 15.72 & 7.31 & 20.53 & 43.88 \\
& Number & 54.98 & 30.31 & 15.79 & 7.57  & 20.69 & 43.83\\
& Question & 55.03 & 30.46 & 16.20 & 7.74 & 20.71 & 44.039 \\
\hline
\multirow{3}{4em}{OQPL} &
Artificial & \textbf{55.60} & 31.03 & 16.56 & 7.89 &  21.03 & \textbf{44.41}\\
& Number & 55.51 & \textbf{31.17} & \textbf{16.79} & \textbf{8.27}  & \textbf{21.2} & 44.38\\
& Question & 55.28 & 30.81 & 16.55 & 8.21  & 21.11 & 44.27\\
\hline
\end{tabular}
\caption{\label{AQPL-VS-OQPL} Model Finetuning Scores}
\end{table*}

The best performing model is the One Question Per Line (OQPL) model with number delimiters, achieving the highest score for BLEU\_2, BLEU\_3, BLEU\_4 and METEOR. For BLEU\_1 and ROUGE\_L, the One Question Per Line (OQPL) model with artificial delimiters performed the best. 

It is interesting to note, however, that the best OQPL models are on average only 0.6917 points better than their corresponding All Questions Per Line (AQPL) counterparts. We hypothesize that this is because not enough of SQuAD's context paragraphs combined with their questions are long enough to cause the moving context window problem (refer to Section \ref{sec-method-data-qpl}) to occur.

This means that the choice between data formatting (OQPL vs AQPL) only matters marginally, given that the context length does not approach the maximum sequence length of the model.

For further analysis, we also extract post-finetuning features from the generated questions such as question length, paragraph context length, and longest sub-sequence (between the paragraph context and generated question) on the best performing model.

A summary of the finetuning results can be found on Table~\ref{AQPL-VS-OQPL}.

From the initial results and generated questions, we observe the following behaviors:
\begin{itemize}
    \item Some generated questions seem to be simply extracting phrases from the paragraph context, and returning them in question form.
    \item From the 2067 sample generated questions, 19 of which do not end with a ``?'' token. Note that we do not refer to such samples as ``full questions.''
\end{itemize}

From these observations, we perform further analysis on our model and its performance indicators.

\subsection{Evaluating Context-Copying}
From the initial results, we observe that a number of generated questions seem to be simply pulled from the given context, with phrase order reversed. 

In order to quantify how frequent this behavior is present in the model, we calculate the longest common subsequence (LCS) between the generated questions and its corresponding context paragraph. From this analysis, we find that, on average, the model tends to take 6.25 tokens from the context paragraph it was given. 

We observe that in cases where this happens, the generated questions tend to be identification type questions (who/what/when/where), which comprise 91.67\% of the total generated samples. 

We hypothesize that the model learned this mode (context-copying) as its most common generation style because of the frequency of identification type questions in the training dataset. As we suspected, SQuAD contains 88.26\% identification type questions in the training set, which lends empirical evidence to our hypothesis. This frequency caused the model to learn context-copying more than other generation styles during finetuning.

In the future, diversifying the style of question-answer pairs in the training set beyond identification type questions will most likely diversity the generation styles of the model.

\subsection{Failure Modes}
After testing, we observe that 19 samples from the generated question set were non-questions, generated by the model in ``failure mode.'' From the 19 samples, we list down two modes:

\begin{enumerate}
    \item The last 3 words of the generated question keeps on repeating.
    \item The generated question was cut prematurely.
\end{enumerate}

Example generations from the two failure modes can be found on Table \ref{table-non-question-cases}.

\begin{figure*}[ht]
    \centering
    \includegraphics[width=\textwidth]{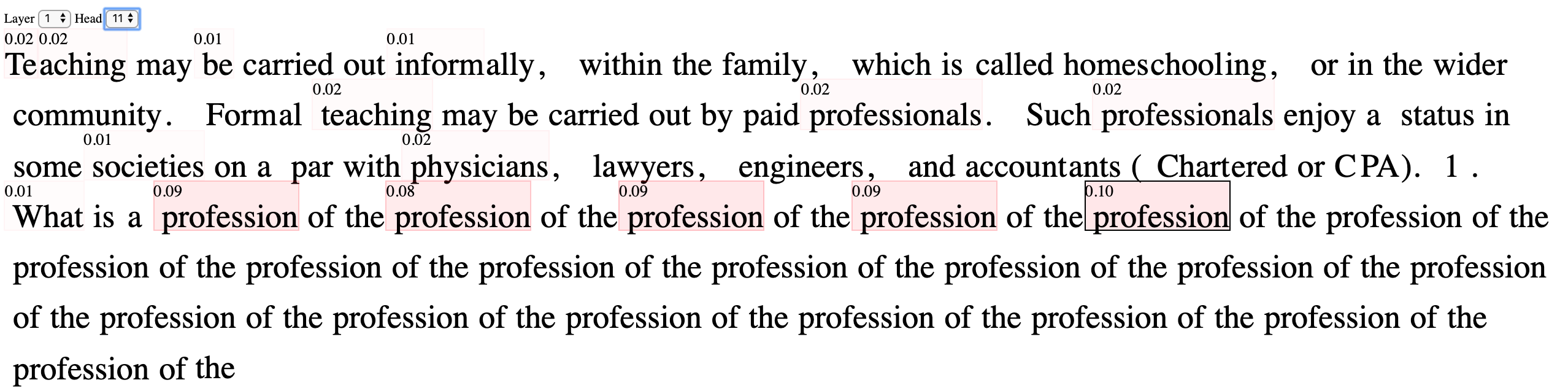}
    \caption{Sample attention visualization for generated outputs of failure mode 1. This example shows the words and the attention values to those words when focusing on the word ``profession,'' which is highlighted in red.}
    \label{fig:attention_confused}
\end{figure*}

\begin{table}[ht]
    \scriptsize
    \centering
    \begin{tabular}{|p{.05\linewidth}|p{.4\linewidth}|p{.4\linewidth}|}
        \hline \textbf{Case} & \textbf{Question} & \textbf{Context} \\ \hline
        1 & What is a profession of the profession of the profession of the profession of the profession of the profession of the profession of the profession of the profession of the profession & Teaching may be carried out informally, within the family, which is called homeschooling, or in the wider community. Formal teaching may be carried out by paid professionals. Such professionals enjoy a status in some societies on a par with physicians, lawyers, engineers, and accountants (Chartered or CPA).\\
        2 & Which newspaper in the United States defined Southern California as including the seven counties of Los Angeles, San Bernardino, Orange, Riverside, San Diego, Ventura and Sant & In 1900, the Los Angeles Times defined southern California as including "the seven counties of Los Angeles, San Bernardino, Orange, Riverside, San Diego, Ventura and Santa Barbara." In 1999, the Times added a newer county—Imperial—to that list.\\
        \hline
    \end{tabular}
    \caption{Examples of failed generations from the best performing model's failure modes.}
    \label{table-non-question-cases}
\end{table}

For failure case 1, where the generated question simply keeps repeating words, we surmise that the attention mechanism is not working properly in pinpointing important context words, which leads to the model being confused in generating the next token.

We look towards visualizing the attention mechanism's behavior while generating for this failure mode. For the following analysis, we point to the attention visualization in Figure \ref{fig:attention_confused}.

When observing the attention scores over the context paragraph for failure case 1, we show that the attention mechanism is ``confused.'' Attention is supposed to point to specific positions in the inputs in order to provide context information better. However, in this case, we see that the attention scores are evenly distributed over a number of random positions in the given context paragraph when generating a token after the word ``profession.'' Instead of helping the model output the best next token, attention ends up not helping at all. This behavior can be seen in multiple attention heads. 

For failure case 2, we surmise that the generation is cut simply because it reached the maximum generation length while copying text from the context, as a consequence of the model's context-copy mode (which it learned as its most common generation mechanism).

\subsection{Optimal Context Length}
\label{sec-sentence-reduction}

In order to understand the limits of the model's robustness, we also look at varying the length of the context paragraph, which we surmise is a performance indicator for the model.

For every context paragraph in the test set with at least 30 sentences, we perform the following:
\begin{enumerate}
	\item The context is fed to the model to generate outputs.
	\item The outputs are scored via BLEU, the results are logged.
	\item We then sentence-split the context paragraph using SpaCy, removing the last sentence, and reconstructing the now-modified context paragraph.
	\item We repeat from step 1 until the modified context paragraph now only has one sentence.
\end{enumerate}

We remove entire sentences instead of reducing the number of words as this interferes with how intact the information is in the context. The model should also be able to produce a question, disregarding performance, even with just one sentence as a context paragraph. We also only test context paragraphs with at most 30 sentences as, on average, this is the most that fit in GPT-2's 1024 maximum sequence length restriction for inputs.

An example of the sentence reduction scheme is shown on Table \ref{sec-sentence-reduction}.

\begin{table}
    \scriptsize
    \centering
    \begin{tabular}{|p{.10\linewidth}|p{.8\linewidth}|}
        \hline \textbf{Sentence Number} & \textbf{Context} \\ \hline
        1 & Proportionality is recognised one of the general principles of European Union law by the European Court of Justice since the 1950s.\\
        2 & Proportionality is recognised one of the general principles of European Union law by the European Court of Justice since the 1950s. \textcolor{blue}{According to the general principle of proportionality the lawfulness of an action depends on whether it was appropriate and necessary to achieve the objectives legitimately pursued.}\\
        3 & Proportionality is recognised one of the general principles of European Union law by the European Court of Justice since the 1950s. \textcolor{blue}{According to the general principle of proportionality the lawfulness of an action depends on whether it was appropriate and necessary to achieve the objectives legitimately pursued.} \textcolor[rgb]{0,0.5,0}{When there is a choice between several appropriate measures the least onerous must be adopted, and any disadvantage caused must not be disproportionate to the aims pursued.}\\
        4 & Proportionality is recognised one of the general principles of European Union law by the European Court of Justice since the 1950s.\textcolor{blue}{According to the general principle of proportionality the lawfulness of an action depends on whether it was appropriate and necessary to achieve the objectives legitimately pursued.}  \textcolor[rgb]{0,0.5,0}{When there is a choice between several appropriate measures the least onerous must be adopted, and any disadvantage caused must not be disproportionate to the aims pursued.}\textcolor{red}{The principle of proportionality is also recognised in Article 5 of the EC Treaty, stating that "any action by the Community shall not go beyond what is necessary to achieve the objectives of this Treaty.}\\
        \hline
    \end{tabular}
    \caption{Sample context paragraph after sentence reduction generation, all of the context in the figure above would be fed to the best performing model. The first sentence, second sentence, third sentence, and fourth sentence  highlighted  in  black, blue,  green, and  red  respectively}
    \label{table-sentence-reduction-sample}
\end{table}

\begin{figure}[ht]
    \centering
    \includegraphics[width=8cm]{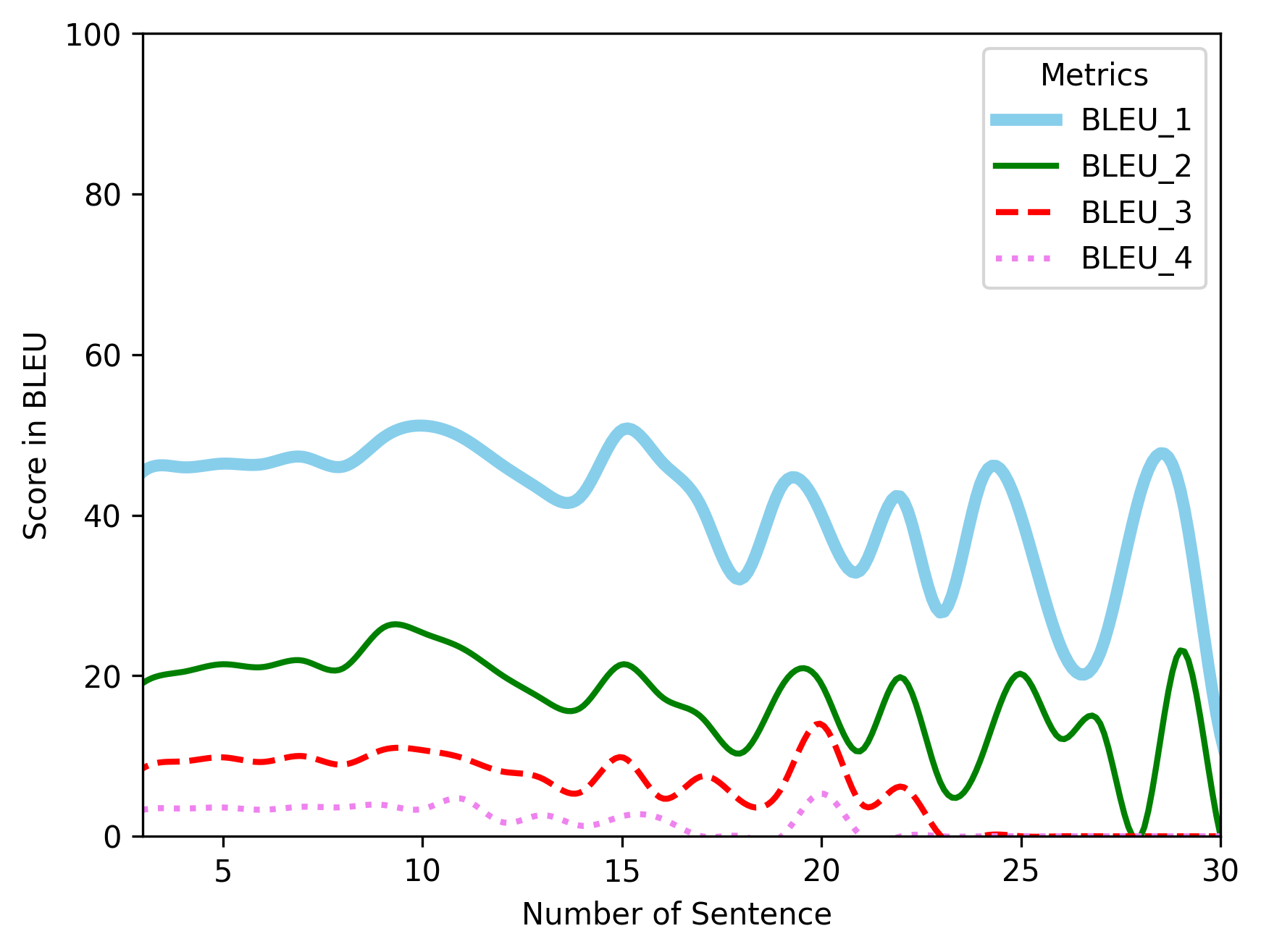}
    \caption{BLEU scores for  each length}
    \label{fig:line_sentence_length_and_score}
\end{figure}

From this analysis, we show that the optimal number of sentences in the context is more or less 10. As the number of sentences increase from 1 to 10, we see that the performance also increases. However, as we increase the number of sentences in the context all the way to 30, the performance is shown to degrade. A graph showing the BLEU scores in relation to the number of sentences in the context paragraph is shown in Figure \ref{fig:line_sentence_length_and_score}.

We hypothesize that this is because the model needs to look at more information in order to identify relevant attention positions as the number of sentences increase. From an interpretative perspective, the performance degradation when the number of sentences increase makes sense because there will be more possible questions to produce from a longer context paragraph than a shorter one. 

A short context paragraph will have a more apparent subject, which can be directly used by the model's context-copy mechanism in order to generate good questions. On the other hand, if the model encounters a long context paragraph where the subject is not apparent (or if the context paragraph has multiple topics/subjects), the context-copy mechanism that the model usually employs will have a hard time pinpointing exact attention positions from where it bases its generated questions from.

Further analyzing the results, we see that BLEU\_1 unsurprisingly degrades the slowest as it only looks at unigram correspondence, while BLEU\_4 degrades the fastest, reaching a score of 0 as early as the 17 sentence mark.

From this analysis, we learn that a higher number of sentences in the context paragraph will give the model more information to generate a question from, too many sentences will confuse the model and cause its performance to degrade.

\subsection{Answer-Awareness}

Given that a number of well-performing previous studies on question generation use answer-awareness, we also test if our single-transformer method will benefit from this additional feature. Answer-awareness refers to the usage of the answer's position or the answer to the question itself, alongside the context paragraph, as input to the model for question generation.

\begin{figure}[ht]
    \scriptsize
    \centering
    \begin{tabular}{|p{.9\linewidth}|}
        \hline
        Super Bowl 50 was an American football game to determine the champion of the National Football League (NFL) for the 2015 season. The American Football Conference (AFC) champion \textcolor{red}{\texttt{[ANSS]} Denver Broncos \texttt{[ANSE]}} defeated the National Football Conference (NFC) champion Carolina Panthers 24–10 to earn their third Super Bowl title. The game was played on February 7, 2016, at Levi's Stadium in the San Francisco Bay Area at Santa Clara, California. As this was the 50th Super Bowl, the league emphasized the "golden anniversary" with various gold-themed initiatives, as well as temporarily suspending the tradition of naming each Super Bowl game with Roman numerals (under which the game would have been known as "Super Bowl L"), so that the logo could prominently feature the Arabic numerals 50. \texttt{[SEP]} Which NFL team represented the AFC at Super Bowl 50? \\
        \hline
    \end{tabular}
    \caption{A sample training example for answer-aware question generation training. The marked answer span is highlighted in red. Uses the ARTIFICIAL delimiter and the OQPL format. Text adapted from SQuAD dataset \protect{\cite{rajpurkar-etal-2016-squad}}.}
    \label{fig-aa-context}
\end{figure}

In order to test this, we employ a OQPL artificial-based formatting scheme, marking the start position of the answer within the context with a special answer start (\verb|[ANSS]|) token, and marking the end of the answer with a special answer-end (\verb|[ANSE]|) token.

A sample input context paragraph with answer-awareness tokens can be found in Figure \ref{fig-aa-context}.

We then follow the same finetuning setup as the original OQPL artificial model, evaluating on BLEU and ROUGE\_L scores. A summary of the finetuning results for the answer-aware model can be found on Table \ref{table-answer-aware}.

\begin{table*}
\centering
\begin{tabular}{llllll}
\hline 
\textbf{Model} & \textbf{BLEU\_1} & \textbf{BLEU\_2} & \textbf{BLEU\_3} & \textbf{BLEU\_4} & \textbf{ROUGE\_L} \\ 
\hline
OQPL Standard & 55.60 & 31.03 & 16.56 & 7.89 & 44.41 \\
OQPL Answer-Aware & 36.07 & 18.83 & 10.95 & 6.40 & 39.80 \\
\hline
\end{tabular}
\caption{\label{table-answer-aware} Summary of Answer-Aware finetuning results.}
\end{table*}

From these results, we can see that the answer-aware models perform significantly worse in terms of BLEU score, and marginally worse than the standard OQPL artificial model in terms of ROUGE\_L. 

We surmise that this is because the model has no inherent idea what to do with the answer-awareness information, and unlike true answer-aware models like UniLM \cite{dong2019unified}, no explicit mechanism that puts importance to the answer-awareness is present in the model. While it is possible for the model to inherently learn to attend to the answer information, this is not deterministic. An explicit, separate mechanism to incorporate answer-awareness in order to help the model learn the feature's significance is still important to have. In the end, the model still performs better without answer-awareness.

\section{Related Literature}

The most prevalent technique for question generation studies is the usage of a sequence-to-sequence (Seq2Seq) model \cite{du2017learning,du-cardie-2018-harvesting,zhao2018paragraph,dong2019unified} in addition to a variety of other features and mechanisms. Attention is also a widely used technique, used by works that employ both standard RNN architectures and Transformer models \cite{zhao2018paragraph,dong2019unified}.

Other studies employ widely different techniques such as using a policy gradient for reinforcement learning \cite{yuan2017machine}, various lingustic features \cite{zhou2017Neural}, and answer awareness \cite{zhou2017Neural,yuan2017machine,zhao2018paragraph,du-cardie-2018-harvesting}.

While most of these works produce robust results, they are complex (Seq2Seq naturally using two neural networks instead of one) and use a lot of extra techniques in order to boost performance. Our work, in comparison, simply uses a single model (one transformer) instead of two in a Seq2Seq setup. It also uses a simple finetuning setup, and does not use any extensive modifications or techniques. However, it produces robust results that are on par with the state of the art in question generation.

\begin{table*}
\centering
\begin{tabular}{lllll}
\hline \textbf{Model} & \textbf{Answer} & \textbf{BLEU\_4} & \textbf{METEOR} & \textbf{ROUGE\_L}\\ \hline
Du et al. (2017) \cite{du2017learning} & - & 12.28 & 16.62 & 39.75 \\
Du et al. (2018) \cite{du-cardie-2018-harvesting} &  \checkmark & 15.16 & 19.12 & - \\
Zhao et al. (2018) \cite{zhao2018paragraph} (s2s+a) &  - & 4.8 & 12.52 & 30.11    \\
Zhao et al. (2018)\cite{zhao2018paragraph} (s2s-a-at-mcp-gsa) &  \checkmark  & 16.38 & 20.25 & 44.48  \\
Dong et al. (2019) \cite{dong2019unified} &  \checkmark & 22.12 & 25.06 & 51.07   \\
GPT2 + attention (ours) & - & 8.26 & 21.2 & 44.38\\  
\hline
\end{tabular}
\caption{\label{table-prev-works} Previous Works with Paragraph Level Input}
\end{table*}

Our model outperforms prior RNN-based Seq2Seq works \cite{du2017learning,du-cardie-2018-harvesting,zhao2018paragraph} in terms of METEOR and ROUGE\_L score. It is worth noting that, in addition to a more complex model setup, \cite{zhao2018paragraph} uses other techniques such as a maxout pointer mechanism and gated self attention mechanisms. Other previous work also use answer awareness, using the positions of the answers in the paragraph, or the answers themselves, as additional features for the model. Our transformer uses none of these extra features, yet still achieves robust METEOR and ROUGE\_L scores that outperform these studies.

Our model performs worse in terms of BLEU\_4 and ROUGE\_L, and slightly worse in terms of METEOR when compared with the recent UniLM work of \cite{dong2019unified}. It is important to note that \cite{dong2019unified} is also the only other work that uses a Transformer for their question generation model. Their incorporation of an answer-awareness mechanism, in addition to the multiple modes of finetuning on a Seq2Seq transformer produces the best results in recent literature.

While our model performs worse than UniLM, we note that UniLM uses a Seq2Seq-based approach, necessitating the use of two separate Transformers: an encoder and a decoder. In contrast, our model relies only on a single Transformer-decoder-based language model, effectively halving model complexity. In addition, our model does not require any sort of answer tagging, making it suitable for situations where this information is not available in the input context. Our model is smaller, less complex, and faster to operate, making it an ideal alternative for a variety of use cases related to question generation.

\section{Conclusion}
\label{sec-conclusion}

Previous attempts at paragraph-level question generation have relied on several additional features and techniques in order to produce state-of-the-art results. In this paper, we demonstrate that a simple single Transformer-based question generation model is able to outperform more complex Seq2Seq methods without the need for additional features, techniques, and training steps. For future work, we plan to evaluate performance on more difficult datasets that pose ``why'' or ``how'' questions as opposed to SQuAD's factoid-only questions. We also look towards training with larger model sizes and evaluating the cost-benefit of using larger models as opposed to more efficient ones.

\bibliographystyle{splncs04}
\bibliography{bibliography.bib}

\end{document}